\documentclass[10pt,twocolumn,letterpaper]{article}
\usepackage{cvpr}
\usepackage{times}
\usepackage{epsfig}
\usepackage{graphicx}
\usepackage{amsmath}
\usepackage{amssymb}
\usepackage{algorithm}
\usepackage{algorithmic}
\usepackage[breaklinks=true,bookmarks=false]{hyperref}
\cvprfinalcopy 

\begin{document}
\title{The Least Restriction for Offline Reinforcement Learning}
\author{Zizhou Su\\
Beijing, P. R. China\\
{\tt suzizhou@foxmail.com} }
\maketitle

\begin{abstract}
Many practical applications of reinforcement learning (RL) constrain the agent to learn from a fixed offline dataset of logged interactions, which has already been gathered, without offering further possibility for data collection. However, commonly used off-policy RL algorithms, such as the Deep Q Network and the Deep Deterministic Policy Gradient, are incapable of learning without data correlated to the distribution under the current policy, making them ineffective for this offline setting. As the first step towards useful offline RL algorithms, we analysis the reason of instability in standard off-policy RL algorithms. It is due to the bootstrapping error. The key to avoiding this error, is ensuring that the agent's action space does not go out of the fixed offline dataset. Based on our consideration, a creative offline RL framework, the Least Restriction (LR), is proposed in this paper. The LR regards selecting an action as taking a sample from the probability distribution. It merely set a little limit for action selection, which not only avoid the action being out of the offline dataset but also remove all the unreasonable restrictions in earlier approaches (e.g. Batch-Constrained Deep Q-Learning). In the further, we will demonstrate that the LR, is able to learn robustly from different offline datasets, including random and suboptimal demonstrations, on a range of practical control tasks. 
\end{abstract}

\section{Introduction}
One of the main reasons behind the success of deep supervised learning \cite{24} is the availability of large and diverse datasets such as the ImageNet \cite{8} to train expressive deep neural networks. By contrast, almost all the RL algorithms assume that every agent has to interacts with an online environment (i.e. a real world environment or an artificial simulator environment) \cite{11}. In this way, the agent collects its own experience for training the actor network and the critic network. Unfortunately, active data collection in the real world (autonomous driving \cite{12}, healthcare \cite{17}, etc.) would be expensive and unsafe. Moreover, building a high-fidelity simulator is not easy too.

Offline RL \cite{11, 12} concerns the problem of learning a policy from a fixed dataset of trajectories, without any further interactions with the environment. This setting could leverage the vast amount of existing logged interactions for real world decision-making problems, like robotics \cite{16}, recommender systems \cite{14}, and dialogues \cite{7}. The effective use of such datasets would not only make the real world RL more practical, but would also enable better generation by incorporating diverse prior experience.

In offline RL, an agent does not receive any new corrective feedback from the online environment. And the agent needs to generalize from a fixed dataset to new online environment during evaluation. In principle, off-policy RL algorithms could learn from data collected by any (unknown) policy. Nonetheless, recent work \cite{22,23} proposes a discouraging view that standard off-policy deep RL algorithms diverge or otherwise yield poor performance in the offline setting.

The rest of this paper is organized as the following: Section 2 explains the background of the offline RL. Section 3 analyzes the reason of why the online algorithms failing in the offline dataset. Then, Section 4 describes a creative offline RL framework, the LR, in details. Finally, Section 5 has a summary discussion.

\section{Background}
At this section, some necessary prior knowledge for offline RL will be introduced.

\subsection{Online Reinforcement Learning}
An interactive environment in RL is typically modeled as a Markov Decision Process (MDP) \cite{20} $< \mathcal{S}, \mathcal{A}, R, P, \gamma >$, where $\mathcal{S}$ is the state space, $\mathcal{A}$ is the action space, $R(s,a)$ is the reward function, $P(s'|s,a)$ is the transition distribution, and $\gamma \in [0,1]$ is the discount factor. A stochastic policy $ \pi(\cdot|s)$ maps each state $s \in \mathcal{S}$ to a probability distribution (density) over actions.

For an agent following the policy $\pi$, the action-value function, denoted $Q^\pi (s,a)$, is defined as the expectation of cumulative discounted future rewards, i.e.:
\begin{equation} Q^\pi (s,a):=\mathbb{E} [\sum_{t=0}^{\infty} \gamma^t R(s_t,a_t) ] \end{equation}
$$ s_0=s, a_0=a, s_t \sim P(\cdot|s_{t-1}, a_{t-1} ), a_t \sim \pi(\cdot|s_t) $$

The goal of RL is to find an optimal policy $\pi^*$ that attains maximum expected return, for which $Q^{\pi ^*}(s,a) \geq Q^\pi (s,a)$ for all $\pi,s,a$. The Bellman optimality equations characterize the optimal policy in terms of the optimal Q-values, denoted $Q^*=Q^{\pi ^*}$, via:
\begin{equation} Q^* (s,a)=\mathbb{E} [R(s,a)] + \gamma \mathbb{E}_{s' \sim P} \text{max}_{a' \in \mathcal{A} } Q^*(s',a')  \end{equation}

The optimal policy $\pi^*$ could be obtained by the Q-Learning algorithm \cite{19}, via iterating the Bellman optimal operator $\mathcal{T}$, defined as:
\begin{equation} (\mathcal{T} \hat{Q} )(s,a) := R(s,a) + \gamma \mathbb{E}_{s' \sim P} [\text{max}_{a'} \hat{Q}(s',a') ] \end{equation}

For large and complex state spaces, the Q-values can be approximated by using neural networks, e.g. the Deep Q Network (DQN) \cite{10}. The DQN optimizes the Q-values network’s parameters $\theta$ by minimizing the mean squared Bellman error $\mathbb{E}_\nu [(Q-\mathcal{T} \hat{Q})^2]$, where $\nu$ is the state occupancy measure under the behavior policy.

In a continuous action space, the maximization $\text{max}_{a’} Q(s’,a’)$ is generally intractable. In this case, actor-critic methods \cite{6} are commonly used, where action selection is performed through another policy network $\pi(s;\theta_{\pi} )$, called the actor, and updated following the Deterministic Policy Gradient Theorem \cite{8}:
\begin{equation} \theta_\pi \leftarrow \text{argmax}_{\theta_\pi} \mathbb{E} [Q (s, \pi(s;\theta_\pi); \theta_Q ) ] \end{equation}
which corresponds to learning an approximation to the maximum of $Q(s,a;\theta_Q)$, by propagating the gradient through both $\pi$ and $Q$. When combined with the DQN to learn $Q(s,a;\theta_Q)$, this algorithm is referred to as the Deep Deterministic Policy Gradient (DDPG) \cite{8}.

\subsection{Offline Reinforcement Learning}
Modern off-policy deep RL algorithms (as discussed above) perform remarkably well on common benchmarks, such as the Atari 2600 Games \cite{18} and the continuous control MuJoCo tasks \cite{9}. Such off-policy RL algorithms are considered “online”, because they alternate between optimizing a policy and using that policy to collect more data. Typically, these algorithms keep a sliding window of most recent experiences in a finite replay buffer, throwing away stale data to incorporate most fresh and “on-policy” experiences.

Offline RL, in contrast to online RL, describes the fully off-policy setting of learning using a fixed dataset of experiences, without any further interaction with the environment. We advocate the use of offline RL to help isolate an RL algorithm’s ability to “exploit” experience and generalize VS. its ability to “explore” effectively. The offline RL setting removes design choices related to the replay buffer and exploration. Therefore, it is easier to experiment and reproduce than the typical online setting.

\begin{figure}
	\centerline{\includegraphics{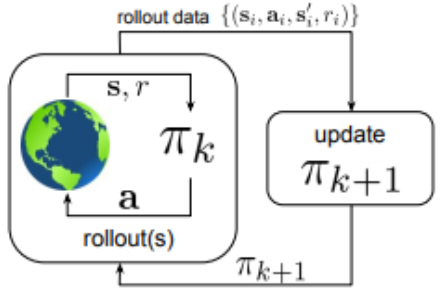} }
	\caption{Online on-policy RL}
\end{figure}

\begin{figure}
	\centerline{\includegraphics{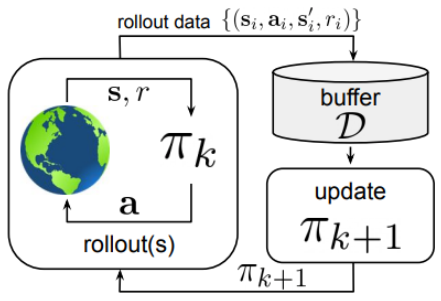} }
	\caption{Online off-policy RL}
\end{figure}

\begin{figure}
	\includegraphics[width=0.6\textwidth]{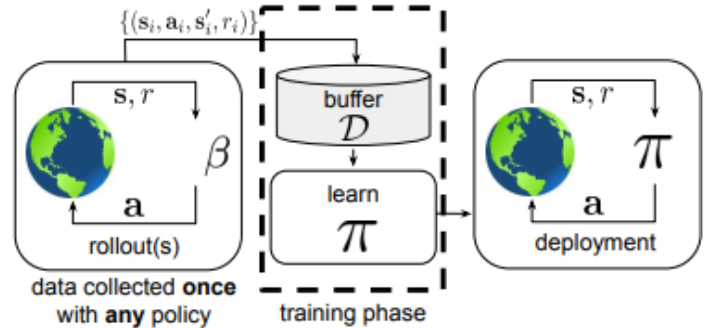}
	\caption{Offline RL}
\end{figure}

Figure 1, 2, 3 illustrate online on-policy RL, online off-policy RL and offline RL, respectively. At on-policy RL, the learned policy $\pi_k$ is updated with streaming data collected by $\pi_k$ itself. At the classic off-policy setting, the agent’s experience is appended to a experience buffer $\mathcal{D}$, and each new policy $\pi_k$ collects additional data, such that $\mathcal{D}$ is composed of samples from $\pi_0$, $\pi_1$,...,$\pi_k$, and all of this data is used to train an updated new policy $\pi_{k+1}$. In contrast, offline RL employs a dataset $\mathcal{D}$ collected by some (potentially unknown) behavior policy $\beta$. The dataset is collected once, and is not altered during training, which makes it feasible to use large previous collected datasets. The training process does not interact with the MDP at all, and the policy is only deployed after being fully trained.  

Offline RL is considered challenging due to the distribution mismatch between the current policy $\pi$ and the offline data collection policy $\beta$, i.e., when the policy being learned takes a different action than the data collection policy, we do not know the reward it would have gotten.

In this situation, the policy constraint methods are the most common methods for offline RL. 
When the action-value function $Q(s,a)$ is being iterated via the following equation:
\begin{equation} Q(s,a) \leftarrow R(s,a) + \gamma Q(s',a') \end{equation}
$$ a' \sim \pi(\cdot|s') $$
These methods ensure that, explicitly or implicitly, the distribution over actions under which we compute the target value, $\pi(\cdot|s')$, is “close” to the collecting dataset behavior distribution $\beta(\cdot|s')$.

For instance, the Batch-Constrained Deep Q-Learning (BCQ) forces $\pi$ being same as $\beta$ via training a Variational Auto-Encoder (VAE) \cite{5} to fit the latent probability distribution in the fixed dataset. Furthermore, the Bootstrapping Error Accumulation Reduction (BEAR) \cite{101} constrains the policy by shrinking the Maximum Mean Discrepancy (MMD) \cite{21} between the unknown behavior policy $\beta$ and the learned policy $\pi$. These constraints are sufficient condition for offline RL, nevertheless are not necessary condition. This reason will be discussed at the next section.

\section{The Key for Offline Reinforcement Learning}
Off-policy RL algorithms, which are with the critic for estimating the action-value function $Q(s,a)$, almost fail to learn on a fixed offline dataset. \cite{100} and \cite{101} demonstrate that, this failure is not caused by the lack of the $<s,a,r,s’>$ transition record, but the error from the $Q(s,a)$ bootstrapping. The source of this instability could be understood by examining the form of the equations for iterating the $Q(s,a)$. Although minimizing the mean squared error corresponds to a supervised regression problem, the targets $Q(s,a)$ for this regression are themselves derived from the current $Q(s',a')$ estimate. The targets $Q(s,a)$ are calculated by maximizing the learned $Q(s',a')$ with respect to the action $a'$ at the next state $s'$. However, the $Q(s,a)$ estimator is only reliable on inputs from the same distribution as its training set.

Have a review on the Equation (2) and (3). For estimating the value of $Q(s,a)$, all the values of $Q(s’,a’), \forall a' \in \mathcal{A}$ must be estimated reliably. And for estimating the value of $Q(s’,a’)$, all the values of $Q(s'',a'')$ must be estimated reliably. And so on and so forth…… Finally, the correct value of $Q(s_{next-to-last}, a_{last} )$ is vitally needed. In the online setting, the agent is easily able to transfer from $<s,a,r,s'>$ to $<s',a',r',s''>$. Then, from $<s',a',r',s''>$ to $<s'',a'',r'',s'''>$. And so on and so forth...... End at the final state $<s_{next-to-last}, a_{last}, r_{last}, s_{final}>$.

Unfortunately, the offline dataset is fixed, without any opportunity to supply new data. Hence, at the offline setting, when transfer from $<s,a,r,s'>$ to $<s',a',r',s''>$, many $a’ \in \mathcal{A}$ are not in this dataset. These actions, which are missing or appear only a little times, are defined as $\textbf{Rare Actions} \ a_{few}$ in this paper.

In other words, at the state $s’$, selecting the action $a’_{few}$, what reward $r'$ will get and which next state $s''$ will arrive, are all unknown, i.e. $<s’,a’_{few},?,?>$. Then, the value of $Q(s’,a’_{few})$ always keep the initial random value. As a result, naïve maximizing the $Q(s’,a’), \forall a' \in \mathcal{A}$ usually absorb the wrong value. Furthermore, the error spread through the Bellman backup, like Equation (2). Naturally, based on the critic with wrong $Q(s,a)$ (see Equation (4) ), the actor could not perform optimally.

Overall, the key for offline RL is that preventing the Rear Actions $a_{few}$ interfering the action-value $Q(s,a)$ backups. Previous works \cite{103}, \cite{13}, \cite{18} explicitly constrain the learned policy $\pi$ to be not far from the behavior policy $\beta$, similarly to behavior cloning. While this is enough to ensure that actions lie in the fixed dataset with high probability, it is overly restrictions. For example, if the behavior policy is close to uniform distribution, the learned policy will behave randomly too, resulting in poor performance, even the data is quite sufficient. The next section will propose a more flexible frameworks for offline RL, though.

\section{The Least Restriction for Offline Reinforcement Learning}
At standard online RL, there is an interesting phenomenon that, the off-policy algorithms \cite{1,2,3}, whose behavior policy (also called exploration policy) $\beta$ is different from the learned policy (also called target policy) $\pi$. But they could succeed in training optimal policies. The reason is that $\pi$ is not same as $\beta$ though, they are quite close, only vary in a random noise or $\epsilon$-greedy.

Accordingly, most state-of-the-art offline RL methods force $\pi$ near the $\beta$. The earliest proposed offline RL method BCQ trains a VAE to simulate the $\beta$ distribution. When the action-value function $Q(s,a)$ is updated, the action selection depends on this VAE.

Besides straightly make $\pi$ close to $\beta$, later researchers try to restrict the “distance” between the $\pi$ distribution and the $\beta$ distribution. Under state $s$, the Rare Action $a_{few}$ one-to-one corresponds to the point, which has a low probability density at $\beta(a|s)$. Avoiding Rare Action means that a learned policy $\pi(a|s)$ has positive density only where the density of the behavior policy $\beta(a|s)$ is more than a threshold, i.e.:
\begin{equation} \forall a,\enspace \beta(a|s) \leq \epsilon \Longrightarrow \pi(a,s)=0 \end{equation}

Based on the above analysis, we propose a creative offline RL framework, the Least Restriction (LR) Framework, which is able to combine with almost all online algorithms.

The LR Framework firstly train a Generative Adversarial Network (GAN) \cite{15} to simulate the dataset collected by $\beta$. For simplicity, the GAN only considers the state-action pair $(s, a)$. After the training completed, the generator of the GAN is very similar with $\beta$ and the discriminator can provide a confidence degree of a $(s, a)$ pair whether belongs to the dataset.

In fact, the offline RL algorithms do not extraordinary diverse from those online. Any online algorithm could turn into the corresponding offline one, via dropping out the procedures of interacting with the environment and iterating the experience buffer.

The extra procedures added to the primary off-policy algorithm by the LR are all at action selection. When the action-value function is being iterated (i.e. $Q(s,a) \leftarrow R + Q(s',a')$), an action $a’$ (under the state $s’$) has been selected by the primary algorithm. As Equation (6), the density of $a’$ at $\beta(a’|s’)$ has to be bigger than the threshold. So the selected $(s’, a’)$ is sent to the discriminator of the GAN, obtaining a confidence degree of this pair. If this degree is below the given threshold, then $a’$ is overlayed by a random noise $\cal{N}$, ($\cal{N}$ is a Gaussian noise and its mean value is zero.) until the confidence degree of $(s’,a’+\cal{N} )$ is bigger than the threshold. Afterwards, the $(s’, a’+\cal{N})$ pair instead of $(s’,a’)$ is used to iterate the $Q(s,a)$. The rest procedures are same as the primary algorithm.

Almost all off-policy online algorithm with the action-value function $Q(s,a)$ (without the policy actor is okay) could combine with our LR Framework. In this section, the DDPG is served as an example. The offline LR-DDPG algorithm is summarized in Algorithm 1.

\begin{algorithm}[htb] 
	\caption{The offline LR-DDPG algorithm} 
	\label{mr}
	{\bf Input:} fixed dataset $\mathcal{D}$, horizon $T$, target network update rate $\tau$, mini-batch size $N$, factor $\gamma$, the threshold $p$, the standard deviation $\sigma$ of the Gaussian noise $\cal{N}$ \\
	\\ {\bf Pretrain:} Train the GAN, getting the discrimitor $Dis$ \\
	\\ Initialize the DDPG's networks: the Q-networks $Q(\cdot|\theta_Q)$, $Q'(\cdot|\theta_{Q'} )$ and the policy networks $\pi(\cdot|\theta_\pi)$, $\pi'(\cdot|\theta_{\pi'} )$ \\
	Set \enspace $\theta_{Q'} \leftarrow \theta_Q$, $\theta_{\pi'} \leftarrow \theta_\pi$ \\
	\\ {\bf for} $t=1$ {\bf to} $T$ {\bf do}
	\begin{algorithmic}
		\STATE{ Randomly sample mini-batch of $N$ transitions }
		\STATE{ $<s,a,r,s'>$ from $\mathcal{D}$ }
		\STATE{ Update the Q-network: }
		\STATE{Set $a'= \pi'(s';\theta_{\pi'} ) $}
		\STATE{}
        \STATE{ {\bf while} $Dis(s',a') < p$ {\bf do} }
        \STATE{ \quad $a' \leftarrow a' + \cal{N}$ }
        \STATE{\bf end while}
        \STATE{}
		\STATE{ Set $y_t = r_t + \gamma Q'(s_{t+1},\pi'(s_{t+1};\theta_{\pi' } ); \theta_{Q'} )$ }
		\STATE{ Update $\theta_Q$ by minimizing the loss function: }
		\STATE{ $L_Q = \frac{1}{N} \sum_t [y_t - Q(s_t, a_t;\theta_Q) ]^2 $ }
		\STATE{}
		\STATE{ Update the policy network using the sampled gradient: }
		\STATE{ $\nabla_{\theta_\pi} \pi | _{s_t} \approx \frac{1}{N} \sum_t \nabla_a Q(s,a;\theta_Q)|_{s=s_t, a=\pi(s_t)} \nabla_{\theta_\pi} \pi(s;\theta_\pi)|_{s_t} $ \\}
		\STATE{}
		\STATE{ Update the target networks: }
		\STATE{ $\theta_{Q'} \leftarrow \tau\theta_{Q'} + (1 - \tau)\theta_Q $ }
		\STATE{ $\theta_{\pi'} \leftarrow \tau\theta_{\pi'} + (1 - \tau)\theta_\pi $ }
	\end{algorithmic}
	\bf end for
\end{algorithm}

\section{Discussion}
The goal in our work is to study offline reinforcement learning with fixed datasets. We firstly analyze how error propagates in standard off-policy RL algorithms. It is due to the use of Rare Actions for computing the target values in the Bellman backup. This naturally leads to that the key for offline RL is avoiding selecting a Rare Action. Armed with this insight, we develop a framework for mitigating the effect of Rare Actions, which we call MR. The LR constrains the backup to use actions that have non-negligible support under the data distribution, but without being overly strict in constraining the learned policy. This LR Framework perfectly keep the balance between the training convergence and the optimal learning. The creative LR Framework is able to combine with almost all off-policy RL algorithms with the action-value function $Q(s,a)$ (without the policy actor is okay). Hence, the novel framework proposed in this paper has a significant superiority that it could conveniently transform the advanced online RL algorithm just comes up to the offline one. This advantage may help the offline RL develop more quickly by the way of absorbing the online RL algorithms.     
\\
\\

\end{document}